\documentclass[sigconf,screen=true,anonymous=False,bookmarks=false,9pt]{acmart}
\usepackage{varwidth}
\usepackage{tcolorbox}

\renewcommand\footnotetextcopyrightpermission[1]{} 
\usepackage{lipsum}
\usepackage{titlesec}

\usepackage{graphicx}
\usepackage{amsmath}
\usepackage{mathtools}
\usepackage{comment}
\usepackage[subrefformat=parens,labelformat=parens]{subfig}
\usepackage{bm}
\usepackage{multirow}
\usepackage{threeparttable,booktabs}
\usepackage{blkarray}
\usepackage{tikz}
\usepackage{balance}
\usepackage{courier}                                       
\usepackage{cleveref}                                      
\usepackage[mathcal]{eucal}
\usepackage[noend]{algpseudocode}
\algrenewcommand\textproc{\texttt}
\makeatletter\let\float@addtolists\relax\makeatother
\usepackage{algorithm}

\DeclarePairedDelimiter\floor{\lfloor}{\rfloor}
\usepackage{filecontents}                                  
\usepackage{pgfplots}
\usepackage{pgfplotstable}
\pgfplotsset{compat=newest}
\usepackage[figuresright]{rotating}

\renewcommand{\vec}[1]{\boldsymbol{#1}}
\newcommand{\vecf}[1]{\boldsymbol{\mathcal{#1}}}

\theoremstyle{plain}

\theoremstyle{definition}
\newtheorem{mydefinition}{\textbf{Definition}}

\graphicspath{{./figs/}}

\definecolor{NVgreen}{RGB}{118,185,0}
\definecolor{NVblack}{RGB}{0,0,0}
\definecolor{NVlgrey}{RGB}{205,205,205}
\definecolor{NVmgrey}{RGB}{140,140,140}
\definecolor{NVdgrey}{RGB}{94,94,94}

\definecolor{NVemerald}{RGB}{0,133,100}
\definecolor{NVamethyst}{RGB}{93,22,130}
\definecolor{NVintel}{RGB}{0,113,197}
\definecolor{NVgarnet}{RGB}{137,12,88}
\definecolor{NVfluorite}{RGB}{250,194,0}

\newtcolorbox{systembox}{colback=NVemerald!5!white, colframe=NVemerald!50!white, title=System:}
\newtcolorbox{userbox}{colback=NVamethyst!5!white, colframe=NVamethyst!50!white, title=User:}
\newtcolorbox{assistantbox}{colback=NVintel!5!white, colframe=NVintel!50!white, title=Assistant:}

\newcommand\iccad{\texttt{ICCAD13}}
\newcommand\lithobench{\texttt{LithoBench}}

\begin{document}

\title{
   GPU-Accelerated Inverse Lithography Towards High Quality Curvy Mask Generation
}

\author{Haoyu Yang, Haoxing Ren}
\affiliation{%
  \institution{NVIDIA Corp.}
}
\email{{haoyuy,haoxingr}@nvidia.com}

\begin{abstract}
Inverse Lithography Technology (ILT) has emerged as a promising solution for photo mask design and optimization.
Relying on multi-beam mask writers, ILT enables the creation of free-form curvilinear mask shapes that enhance printed wafer image quality and process window. 
However, a major challenge in implementing curvilinear ILT for large-scale production is mask rule checking, an area currently under development by foundries and EDA vendors. 
Although recent research has incorporated mask complexity into the optimization process, much of it focuses on reducing e-beam shots, which does not align with the goals of curvilinear ILT. 
In this paper, we introduce a GPU-accelerated ILT algorithm that improves not only contour quality and process window but also the precision of curvilinear mask shapes. 
Our experiments on open benchmarks demonstrate a significant advantage of our algorithm over leading academic ILT engines. Source code will be available at \url{https://github.com/phdyang007/curvyILT}.
\end{abstract}

\maketitle
 \renewcommand{\shortauthors}{Yang et al.}

\section{Introduction}
\label{sec:intro}
Lithography plays a crucial role in semiconductor manufacturing. 
However, a mismatch between lithography technology and the critical dimensions of chips leads to the optical proximity effect, posing challenges to technological advancement. 
To mitigate this issue, chip design photomasks must be optimized to correct for lithography proximity distortion, a process known as mask optimization.

\textbf{Optical proximity correction} (OPC) is the most widely used approach for mask optimization \cite{OPC-DATE2015-Kuang, OPC-JM3-2016-Matsunawa, MEEF-TSM2000-Wong}. 
It involves dividing the edges of chip design polygons into segments, which are then adjusted using heuristic rules to counteract optical proximity effects. 
However, as chip feature sizes continue to shrink, the limited robustness of heuristic optimization demands extensive engineering effort, jeopardizing design turnaround time and production yield.
\textbf{Inverse lithography technologies} (ILT), with their gradient-based free-form optimization, provide a broader solution space that can effectively address critical patterns where traditional OPC falls short. Despite this advantage, ILT has long faced a dilemma: while free-form optimization leads to better convergence, it also presents a significant manufacturing challenge, as mask shops struggle to produce these complex free-form masks efficiently. A common workaround is to approximate the ILT-generated mask with rectangles, aligning the ILT output with OPC shape rules. However, this approach sacrifices some of the optimality that ILT offers.
Recently, a multi-beam mask writer was introduced for advanced lithography mask manufacturing \cite{mbmw}. This innovation maintains a consistent mask production time, enabling the direct fabrication of freeform or curvilinear masks. Though curvilinear ILT faces mask writing challenges to enforce clearance of mask rule check \cite{curvyMRC}, it is now feasible to apply curvilinear inverse lithography technology (ILT) across a significant portion of chip design layers, leading to improved quality of results (QoR).
The comparison among OPC, ILT, and Curvy ILT are listed in \Cref{tab:opc-category}, and the development of Curvy ILT is our focus.

\begin{table}[tb!]
\centering
\caption{Mask optimization solution. Our efforts focus on the direct generation of curvy ILT, with a specifically designed algorithm for better curvature and reduced mask artifacts.}
\label{tab:opc-category}
\def\arraystretch{1.2}
\begin{tabular}{p{0.2\linewidth} | p{0.2\linewidth} | p{0.2\linewidth} | p{0.2\linewidth}}
\toprule
Solution          & OPC                                         & ILT                                         & \textbf{Curvy ILT}                            \\ \midrule
Mask Writer       & Variable Shaped Beam                        & Variable Shaped Beam                        & Multibeam                            \\ \midrule
Mask Rule & Manhattan Geometry Constraints & Manhattan Geometry Constraints & Width, Area, Curvature                \\ \midrule
Efficiency   & Fast      & Slow      & Slow \\ \midrule
Solution Space   &Small      &Medium       &Large \\ \midrule
Optimizer   & Heurestic       & Gradient       &  Gradient \\ \midrule
Example      & \includegraphics[width=.1\textwidth,trim={200 400 200 400},clip]{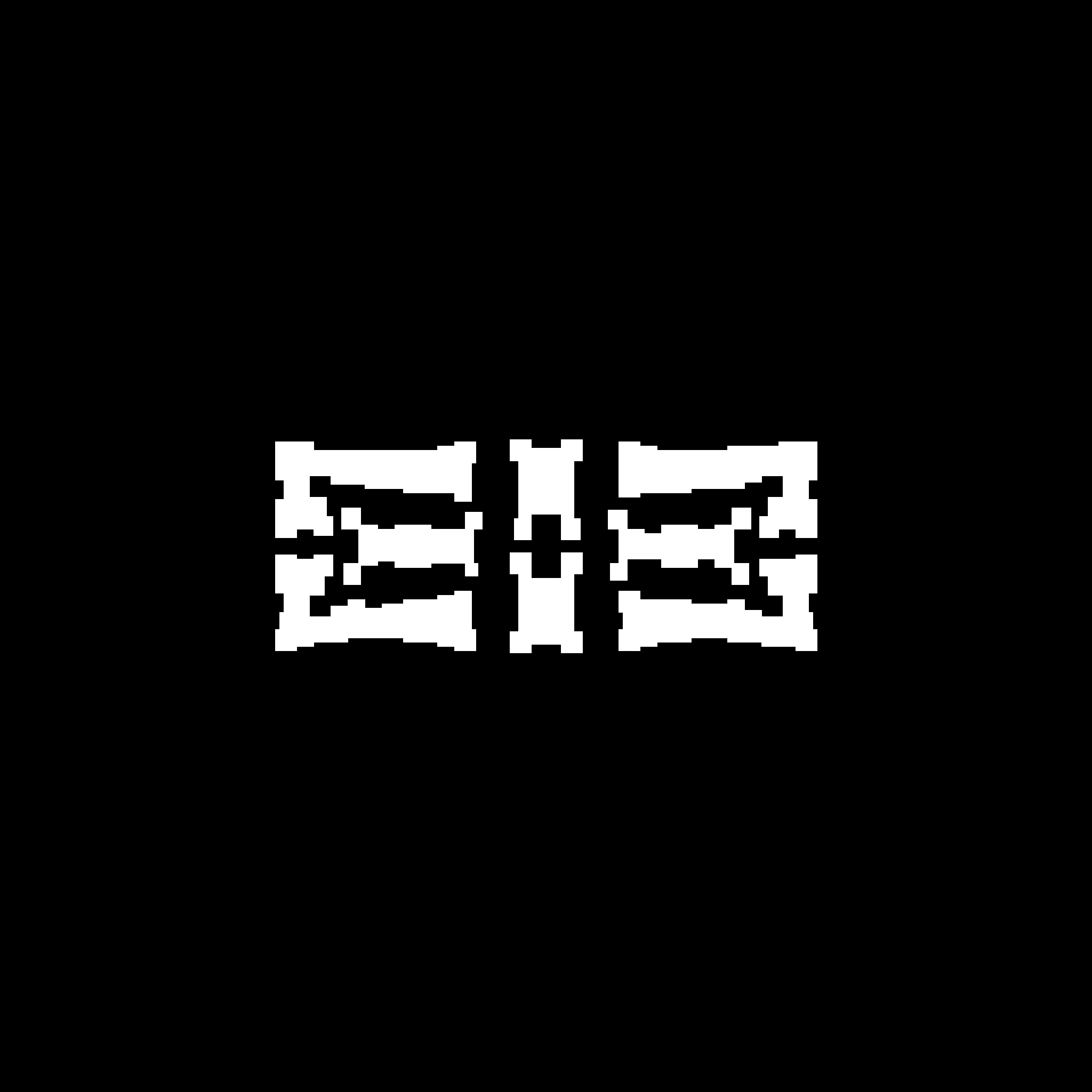} & \includegraphics[width=.1\textwidth,trim={200 400 200 400},clip]{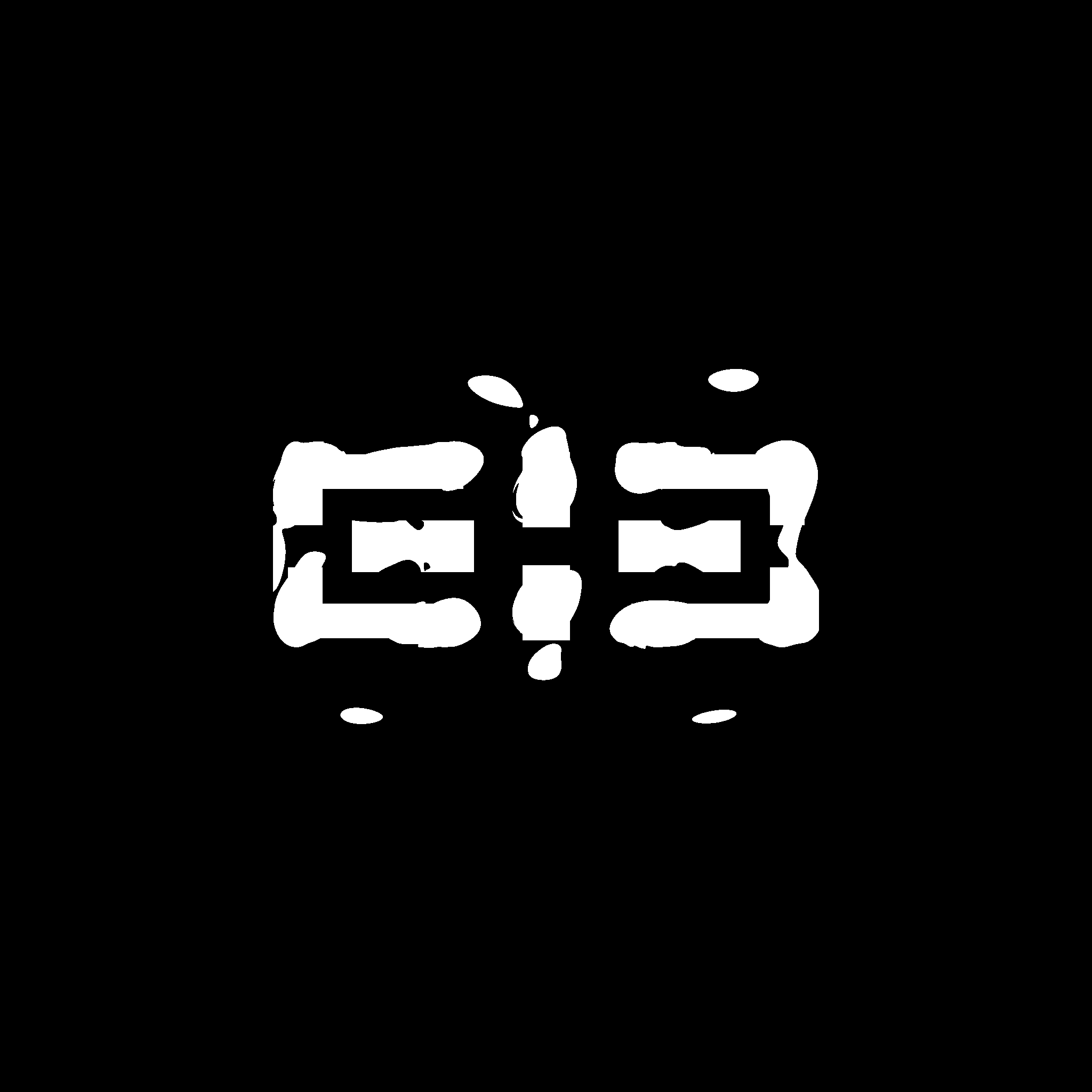}  &\includegraphics[width=.1\textwidth,trim={250 610 150 190},clip]{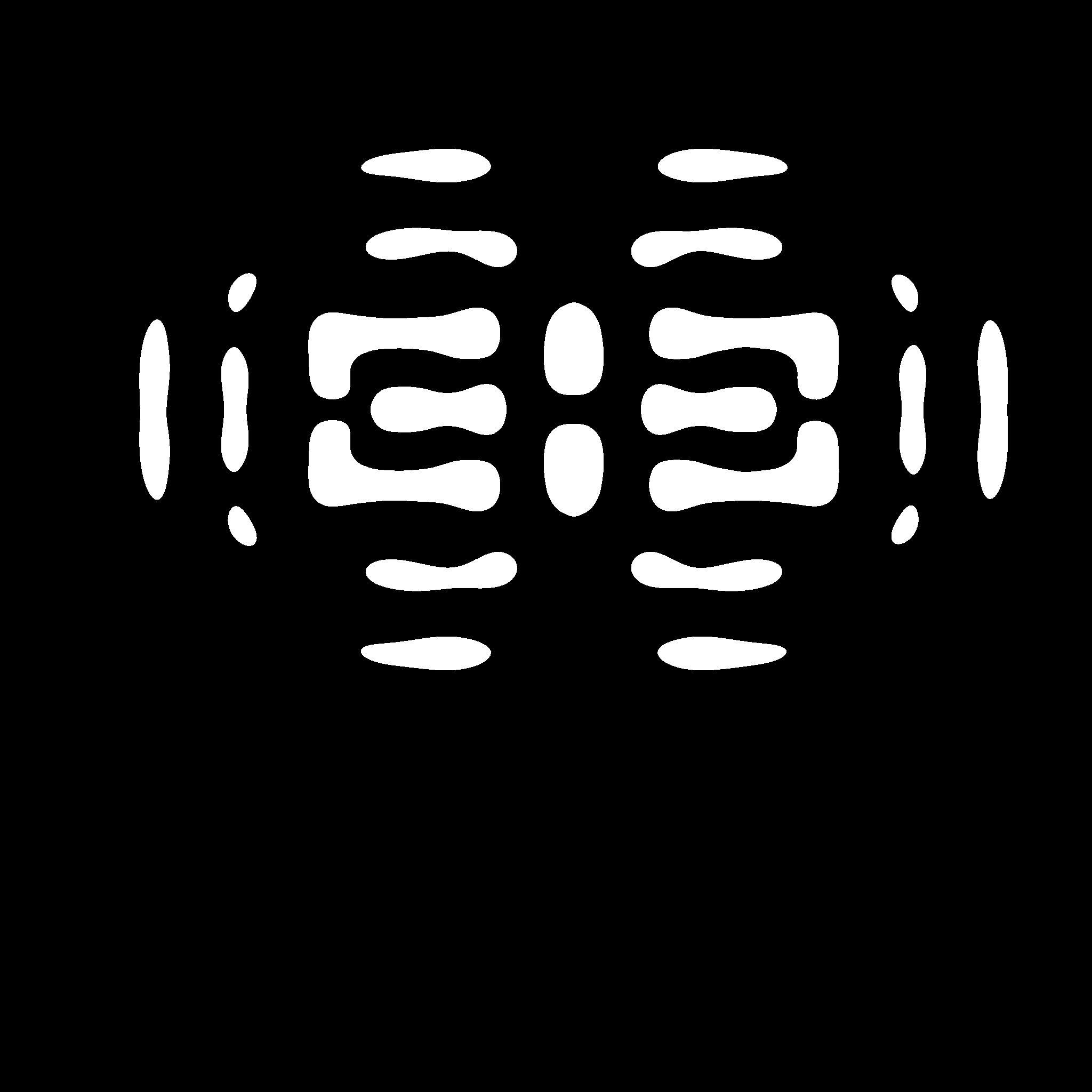}    \\ \bottomrule
\end{tabular}
\end{table}

ILT has garnered significant attention in academic research due to its promising advantages. 
Much of this research has centered on enhancing algorithmic efficiency and optimizing the quality of the final simulated wafer. 
For example, Wang et al. \cite{OPC-DAC2022-Wang} developed A2-ILT, introducing a spatial attention layer to regulate mask gradients, which however fails the growth of sub-resolution assist features (SRAFs)—a critical aspect for process window optimization. 
Additionally, Yu et al. \cite{OPC-DATE2021-Yu} proposed using a level-set function to model the mask, improving smoothness during ILT procedures. More recently, an efficient ILT implementation was presented by Sun et al. \cite{OPC-DAC2023-Sun}, which has become the state-of-the-art by utilizing multi-level lithography simulations at different resolutions to achieve faster convergence.
A similar implementation is also introduced in OpenILT \cite{openilt}.
However, these efforts primarily address the mask manufacturing challenges associated with VSB technology and are not directly applicable to curvilinear mask optimization. 
For instance, there has been little focus on eliminating isolated artifacts that breach shape area constraints, and the balance between quality of results (QoR) and mask smoothness has not been adequately managed \cite{OPC-DAC2023-Sun,OPC-DATE2021-Yu}.
To overcome the limitations of previous work and encourage further research into curvilinear ILT solutions, we introduce a new GPU-accelerated ILT algorithm that: 1) improves upon existing algorithms to achieve better optimality, and 2) addresses the challenges of curvilinear mask writing using differentiable morphological operators.
Our major contributions include:
\begin{itemize}
\item We thoroughly analyze the limitations of existing academic ILT algorithms and have developed a new algorithm that improves optimization convergence and enhances mask quality.
\item We develope the idea of curvilinear design retargeting to allow ILT solvers to optimize toward corner-smoothed targets leading to faster and better convergence. 
\item We introduce a differentiable morphological operator that can be seamlessly integrated into legacy ILT algorithms to control mask curvature and shape without compromising the final quality of results (QoR).
\item We conduct experiments on layers from both real-world and synthetic designs, demonstrating the superior performance of our algorithm.
\end{itemize}

Reminder of the manuscript is organized as follows: 
\Cref{sec:prelim} introduces related works and fundamental terminologies associated with mask optimization and ILT.; 
\Cref{sec:algo} provides a detailed description of the proposed ILT algorithm.;
\Cref{sec:result} presents a comprehensive analysis of the experimental results for our algorithm across various design layers;
and \Cref{sec:conclu} discusses future work and concludes the paper.

\section{Preliminaries}
\label{sec:prelim}
\subsection{Notations}
Throughout the paper, we use lowercase letters for scalar (e.g.~$x$), bold lowercase letters for vector (e.g.~$\vec{x}$) and bold uppercase letters for matrix (e.g.~$\vec{X}$).
Specifically, we use $\vec{X}(i,j)$ for single entry index and $\vec{X}(i_1:i_2,j_1:j_2)$ for block index.
A full list of notations and symbols is shown in \Cref{tab:notation}.

\begin{table}[]
\centering
\caption{Notations and symbols used throughout this paper.}
\label{tab:notation}
\def\arraystretch{1.2}
\begin{tabular}{c|c}
\toprule
Notation & Description \\ \midrule
    $\vec{M}$  & Mask image            \\
    $\vec{M}_c$ & Continuous tone mask image           \\ 
    $\vec{Z}^\ast$& Manhattan design target           \\ 
    $\vec{Z}^\ast_r$ & Retargeted design target           \\ 
    $\vec{X}(i,j)$ & The entry of $\vec{X}$ given $i,j$ index           \\
    $\vec{X}(i_1:i_2,j_1:j_2)$ & A sub-block of $\vec{X}$ given $i_1,i_2,j_1,j_2$ index           \\ 
    $\vec{A} \otimes \vec{B}$ & Convolution of $\vec{A}$ by $\vec{B}$          \\ 
    $\vec{A} \odot \vec{B}$  & Element-wise product between $\vec{A}$ and $\vec{B}$          \\ 
    $\vec{A} \oplus \vec{B}$  & Dilation of $\vec{A}$ by the structuring element $\vec{B}$           \\ 
    $\vec{A} \ominus \vec{B}$  & Erosion of $\vec{A}$ by the structuring element $\vec{B}$          \\
    $\mathcal{F}(\vec{A})$  & The Fourier transform of $\vec{A}$          \\\bottomrule
\end{tabular}
\end{table}

\subsection{Forward and Inverse Lithography}
Forward lithography simulation encompasses the fundamental mathematics of mask optimization, modeling the lithographic process through a series of approximation equations.
In this context, we utilize the widely adopted Hopkin's Diffraction model along with a constant resist threshold, as described below:
\begin{align}
    \vec{I} = \sum_{i=1}^{k} \alpha_i ||\mathcal{F}^{-1}(\vecf{M} \odot \vecf{H}_i)||_2^2,
    \label{eq:aerial}
\end{align}
where $\vecf{M}=\mathcal{F}(\vec{M})$ represents the rasterized mask image $\vec{M}$ in the Fourier domain, $\vecf{H}_i$'s are lithography transmission kernels which are pre-computed given lithography system configurations, and $\vec{I}$ is the aerial image which contains the light intensity projected on the resist material.
Through constant resist threshold modeling, we can derive the final resist image as follows:
\begin{align}
    \vec{Z}(i,j) = 
    \begin{cases}
        1, & \text{if~~} \vec{I}(i,j) \ge D_\text{th},\\
        0, & \text{if~~} \vec{I}(i,j) < D_\text{th}.
    \end{cases}
    \label{eq:resist}
\end{align}
Thus, the ILT problem can be defined as finding the mask $\vec{M}$ such that the resulting resist pattern $\vec{Z}$ closely matches the intended design $\vec{Z}^\ast$. In practical manufacturing, the lithographic system may not operate under ideal conditions, leading to process variations where the resist pattern is either larger ($\vec{Z}_\text{outer}$) or smaller ($\vec{Z}_\text{inner}$) than theoretically predicted. A robust ILT algorithm aims to generate a mask that minimizes these process variations.

Under our assumption, we have $\vec{M},\vec{Z} \in \{0,1\}$. However, to make the mask optimization end-to-end differentiable during ILT, a practice is to convert the binary mask to a continuous tone mask through the sigmoid function,
\begin{align}
    \vec{M}_c = \dfrac{1}{1+\exp[-\beta_1 (\vec{M}-M_s)]},
    \label{eq:cmask}
\end{align}
where $\beta_1$ is called the mask steepness and $M_s$ is a shift parameter and is usually set to 0.5 that yields better SRAF generation \cite{OPC-DAC2023-Sun}.
Similarly, the resist image is also converted to the continuous domain,
\begin{align}
    \vec{Z} = \dfrac{1}{1+\exp[-\beta_2 (\vec{I}-D_\text{th})]}.
    \label{eq:cresist}
\end{align}
Accordingly, ILT can be mathematically formulated as follows:
\begin{align}
    \min_{\vec{M}}~ &f(\vec{M},\vec{Z}^\ast), \label{eq:ilt-obj} \\
    \text{s.t.~} & \text{\Cref{eq:aerial,eq:cmask,eq:cresist}},
\end{align}
where $f$ is some objective function to satisfy ILT QoR requirements.

\subsection{Morphological Operator}
\begin{figure}[tb!]
	\centering
	\includegraphics[width=.1\textwidth]{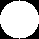}
	\caption{Example disc-shaped structuring element with size 39$\times$39.}
	\label{fig:disc-element}
\end{figure}

\begin{figure}[tb!]
	\centering
	\subfloat[Reference]{\includegraphics[width=.14\textwidth,trim={350 660 250 240},clip]{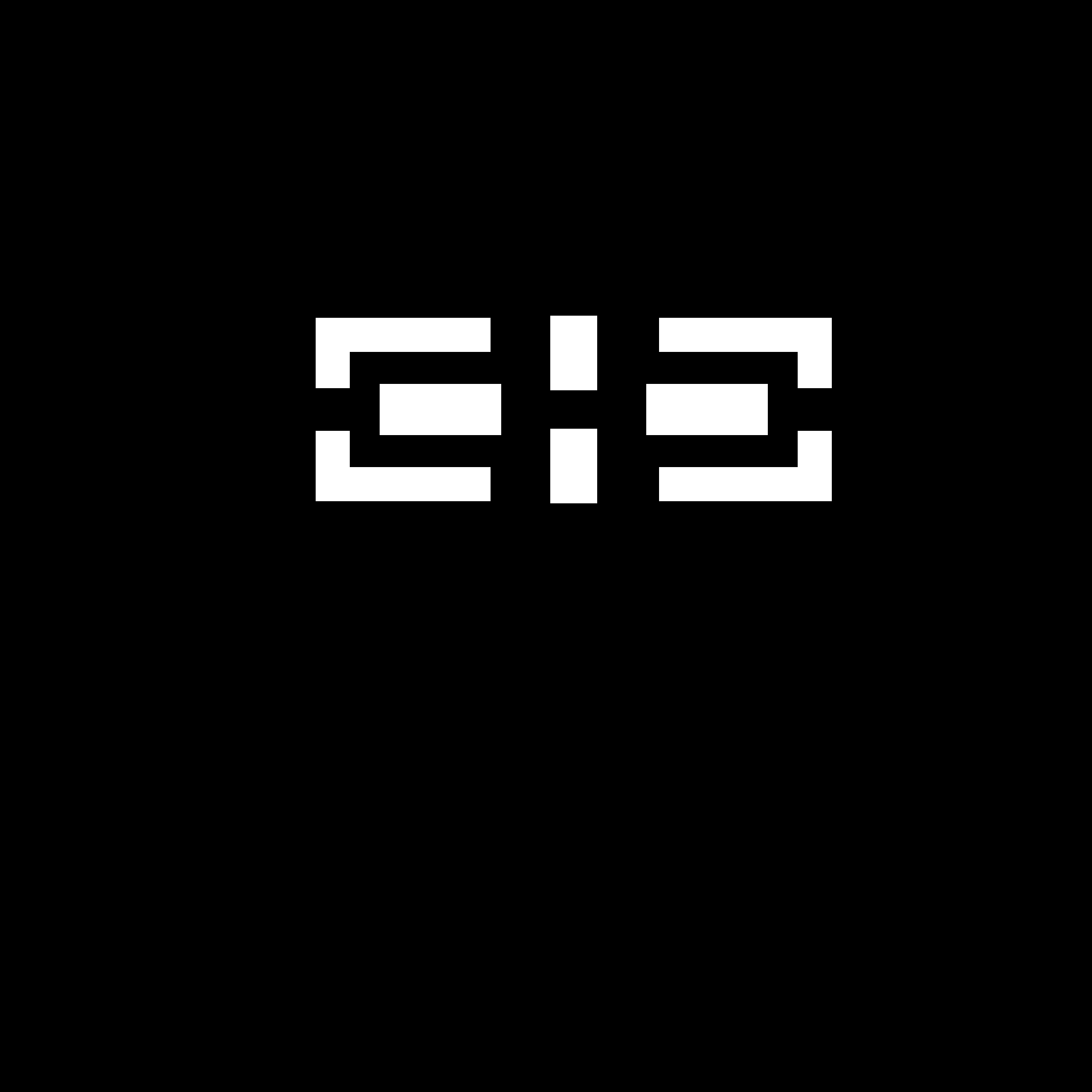}}
	\subfloat[Dilation]{\includegraphics[width=.14\textwidth,trim={350 660 250 240},clip]{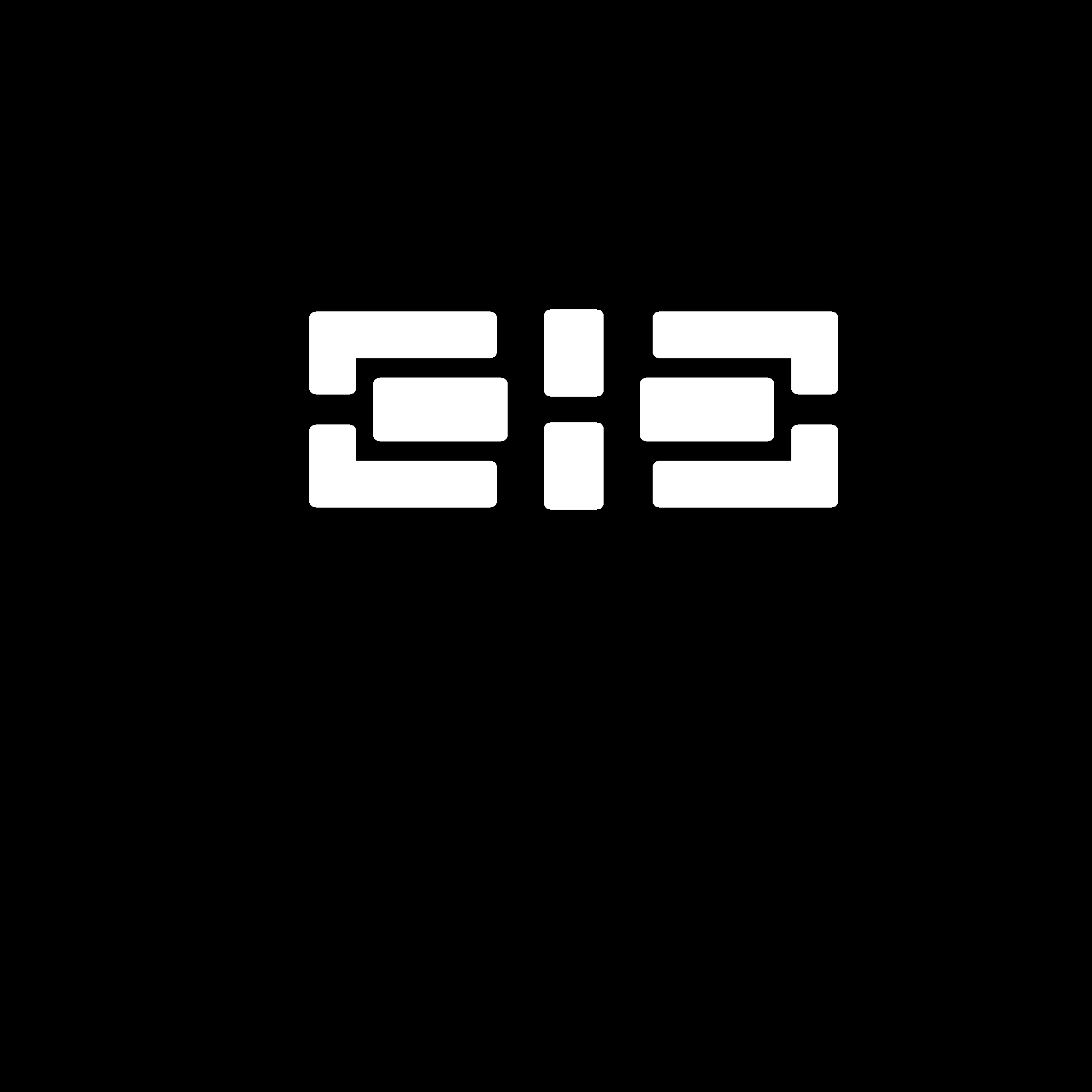}}
	\subfloat[Erosion]{\includegraphics[width=.14\textwidth,trim={350 660 250 240},clip]{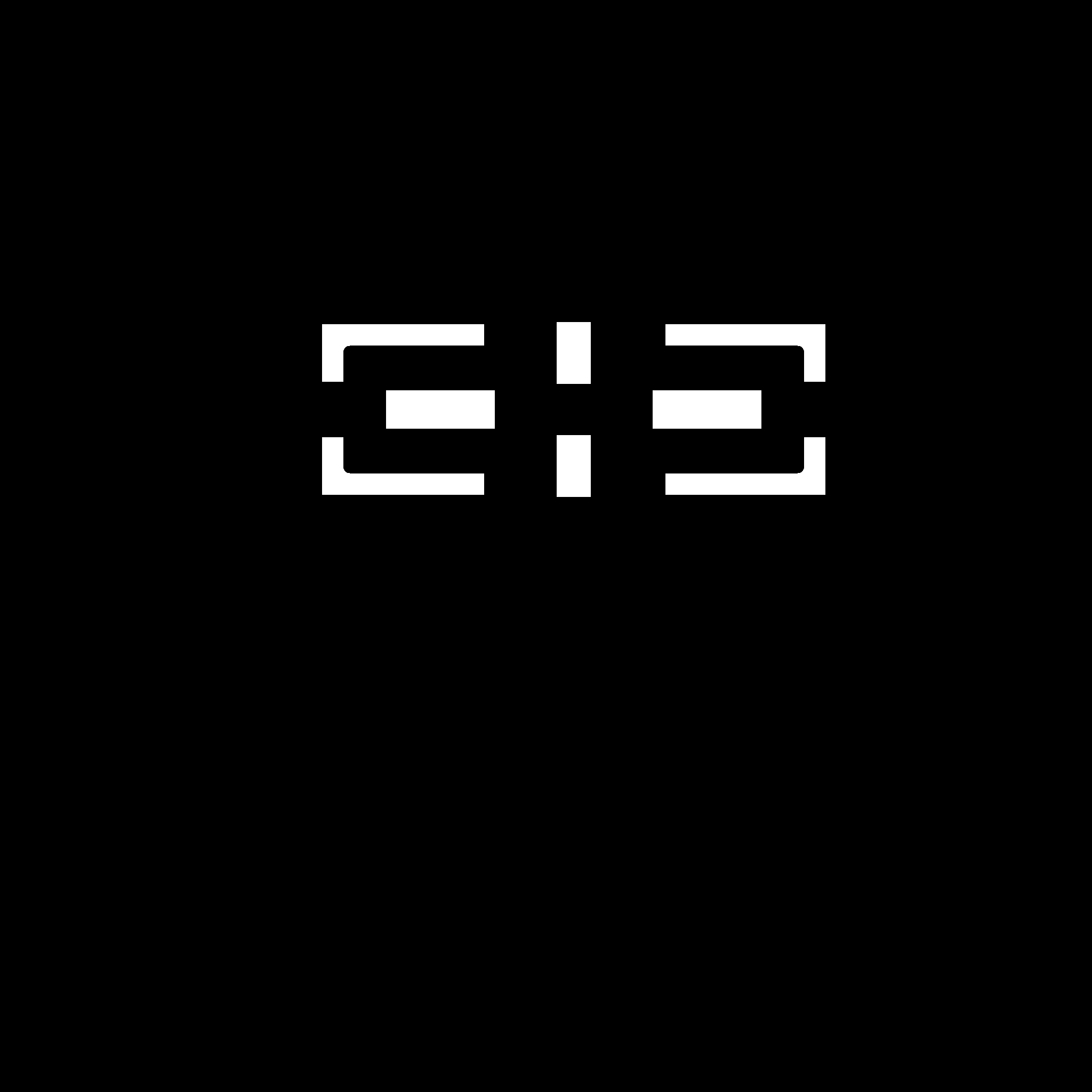}} \\
	\subfloat[Opening]{\includegraphics[width=.14\textwidth,trim={350 660 250 240},clip]{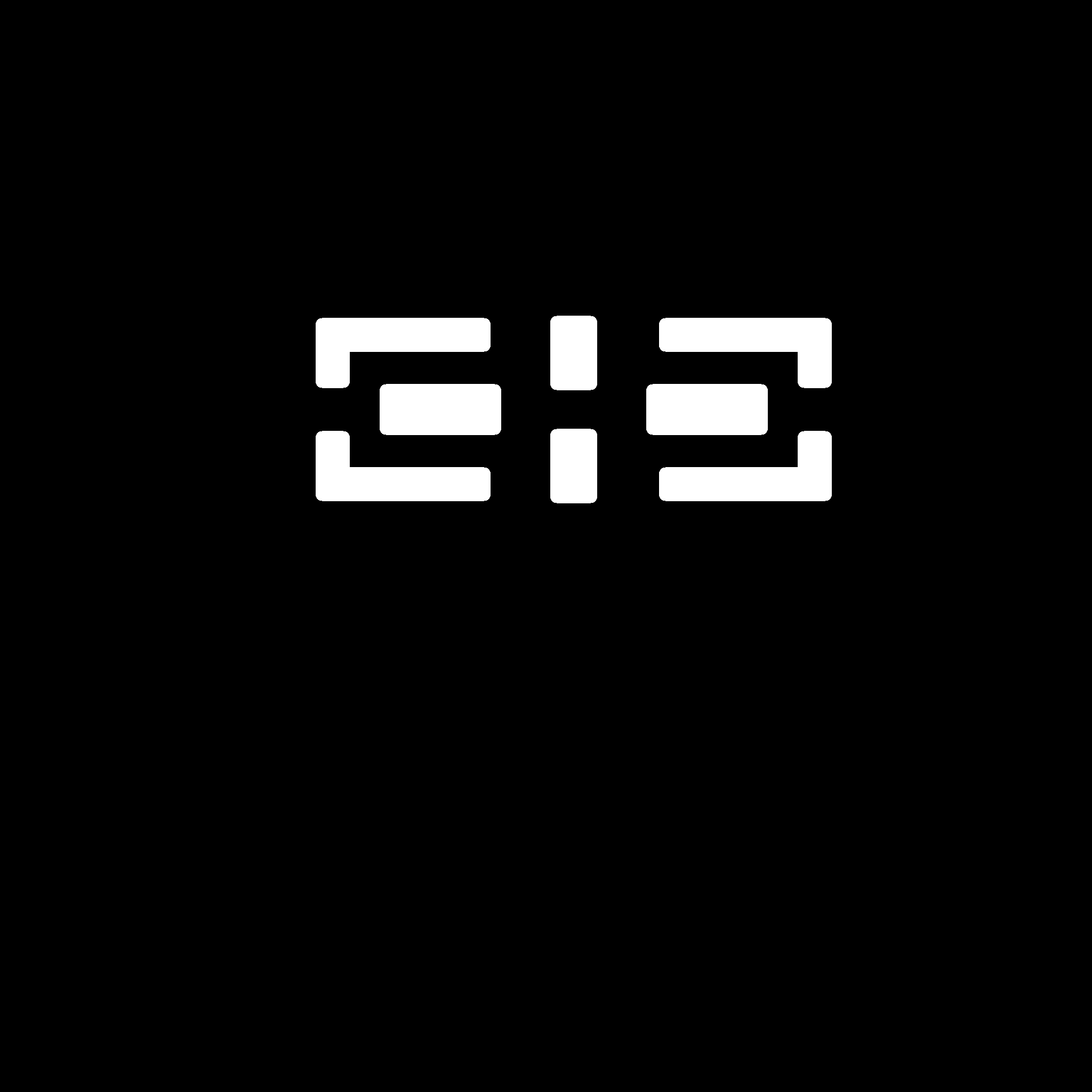}}
	\subfloat[Closing]{\includegraphics[width=.14\textwidth,trim={350 660 250 240},clip]{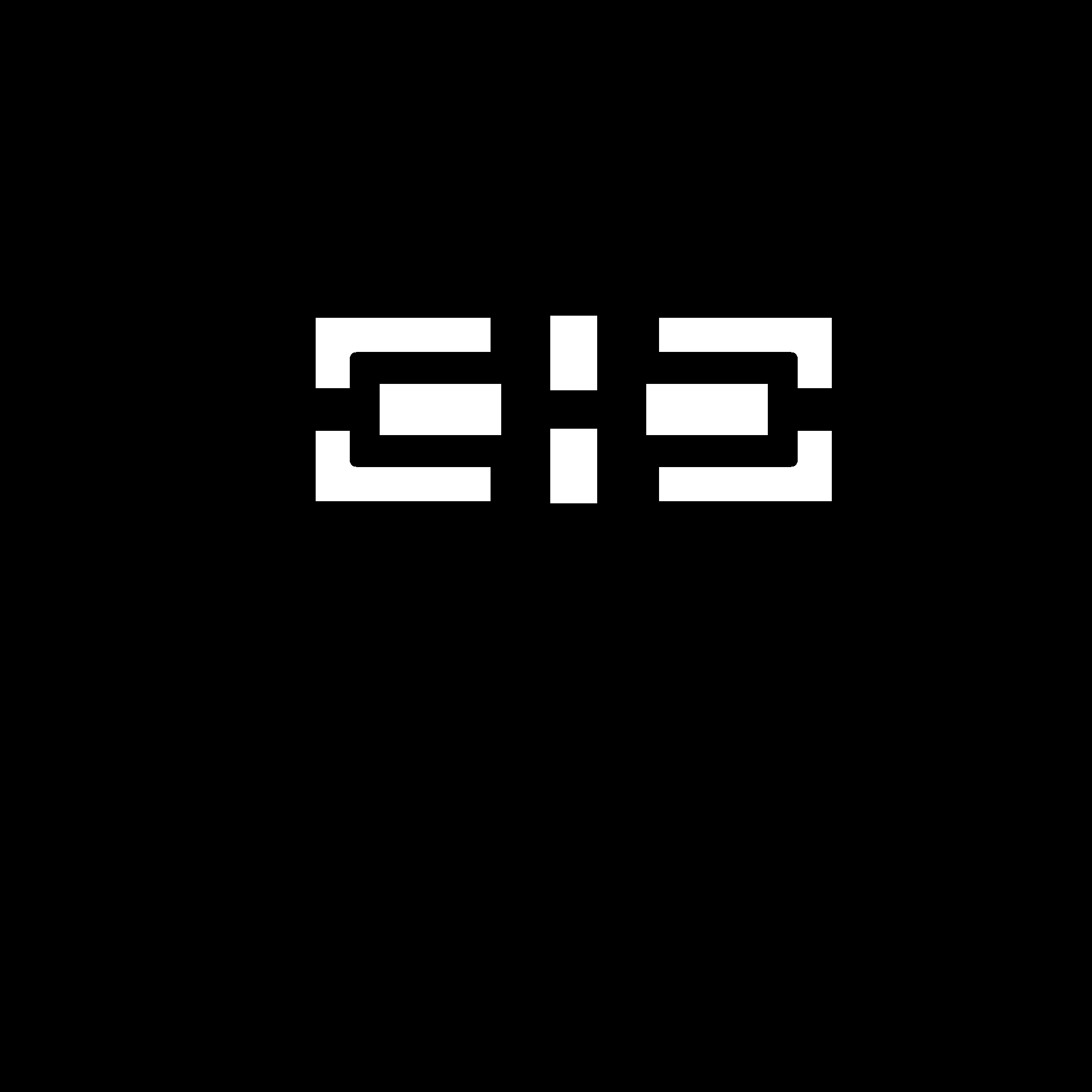}}
	\subfloat[CDR]{\includegraphics[width=.14\textwidth,trim={350 660 250 240},clip]{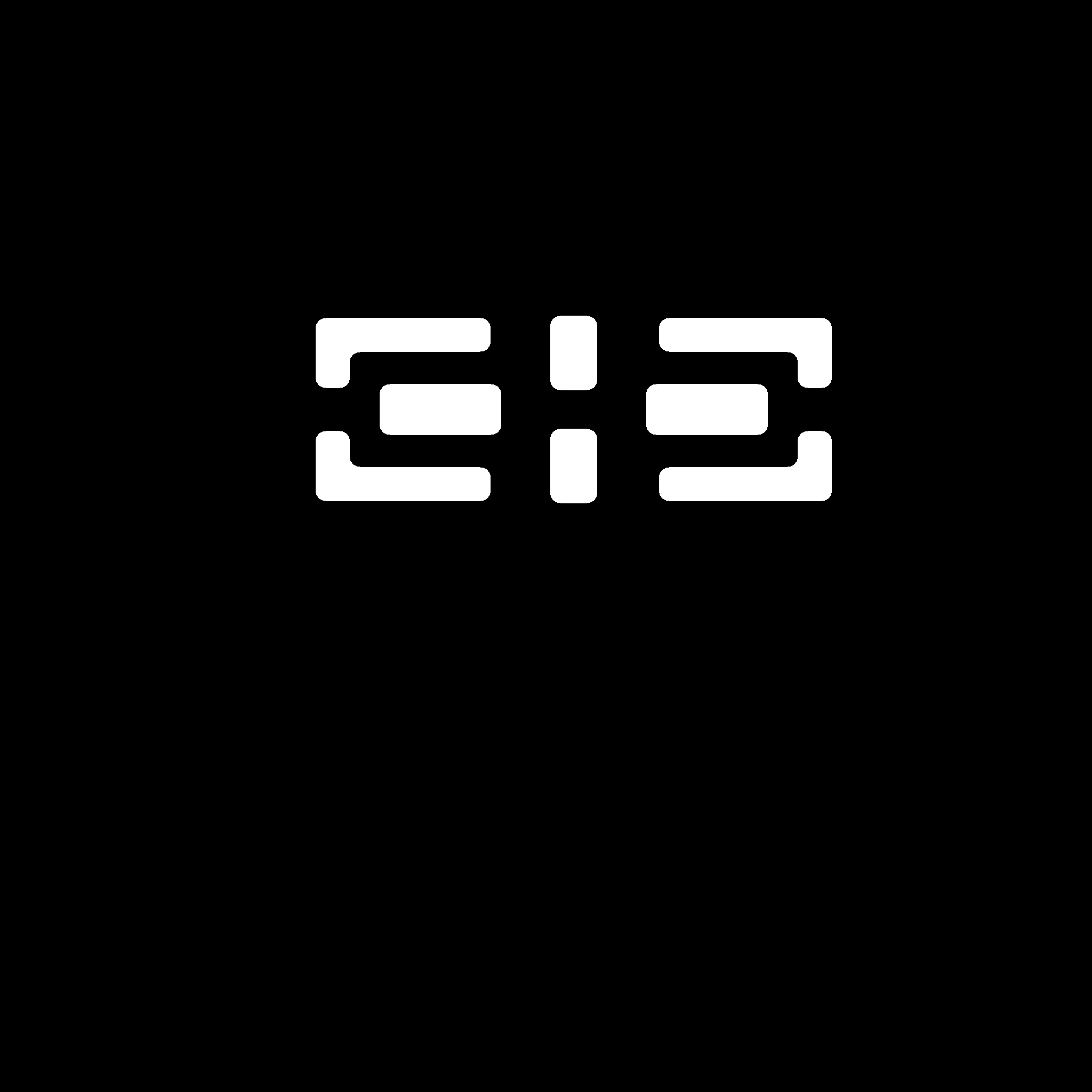} \label{fig:cdr-example}}
	\caption{Visualization of morphological operators with eclipse structural element applied on binary images. (a) The original design image with Manhattan shapes; 
	(b) Dilation enlarges the shapes in the original image;
	(c) Erosion etches the original image that yields smaller shapes;
	(d) Opening rounds the convex corners of each shape;
	(e) Closing rounds the concave corners of each shape;
	(f) CDR rounds both the convex and concave corners of each shape.}
	\label{fig:morph}
\end{figure}

Morphological operator \cite{morphological} started to get noticed in the early 1980s as a theory and technique for processing geometrical structures.
The basic operators include erosion, dilation, opening and closing,
which poses different processing effects on binary geometrical images \footnote{Morphological operators are also applicable on other formats like grayscale images and graphs, which are beyond the scope of this paper.}. 
Following tradition, we use the following notations to represent basic morphological operators:
\begin{align}
	f_e (\vec{A},\vec{B})&=\vec{A}\ominus \vec{B},~~\text{(Erosion)} \\
	f_d (\vec{A},\vec{B})&=\vec{A}\oplus \vec{B},~~\text{(Dilation)} \\
	f_o (\vec{A},\vec{B})&=f_d (\vec{A},\vec{B}) \ominus \vec{B},~~\text{(Opening)}  \label{eq:opening}\\
	f_c (\vec{A},\vec{B})&=f_e (\vec{A},\vec{B}) \oplus \vec{B},~~\text{(Closing)} \label{eq:closing}
\end{align}
where $\vec{A}$ is the input and $B$ is the predefined structuring element.
In this paper we use a disc shape as the structuring element, as exemplified in \Cref{fig:disc-element}.
The effects of four basic morphological operators using eclipse structuring element are depicted in \Cref{fig:morph}, where 
\textit{Dilation} of $\vec{A}$ by $\vec{B}$ results in the locus of $\vec{B}$ when the center of $\vec{B}$ traverses inside $\vec{A}$,
and \textit{Erosion} of $\vec{A}$ by $\vec{B}$ gives the locus of the center of $\vec{B}$ when $\vec{B}$ traverses inside $\vec{A}$.
\textit{Opening} and \textit{Closing} can be derived from combinations of dilation and erosion.
By definition, we can observe several interesting properties of morphological operators:
\begin{enumerate}
	\item Erosion removes shapes smaller than the structuring element.
	\item Dilation merges shapes if their closest distance is smaller than the half diameter of the structuring element.
	\item Opening and Closing do not modify the critical dimension of Manhattan shapes with properly set structuring elements.
\end{enumerate}

\subsection{Evaluation}
\begin{figure}[tb!]
    \centering
    \includegraphics[width=0.85\linewidth]{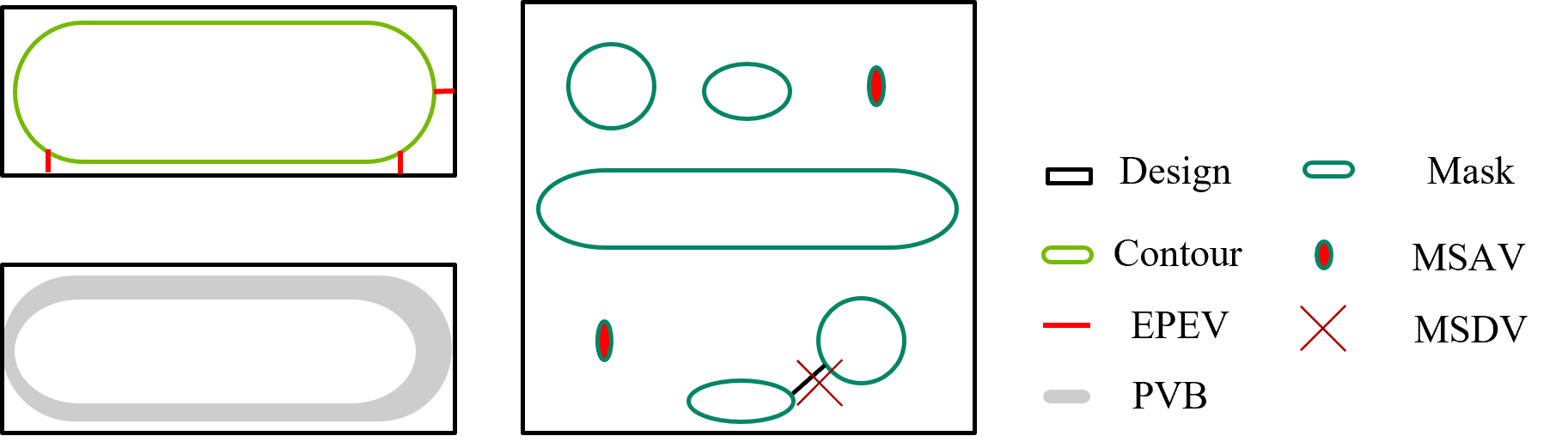}
    \caption{Mask optimization evaluation metrics. EPE violation (EPEV) and PVB are two major measurement in terms of mask lithography quality. We also employ mask shape area violation (MSAV) and mask shape distance violation (MSDV) to represent the curvilinear mask rules.}
    \label{fig:metric}
\end{figure}

We follow the standard mask evaluation metrics to measure the performance of our ILT algorithm,
which include edge placement error violation (EPEV) count and the process variation band area (PVB) (\Cref{fig:metric}).

\begin{mydefinition}[EPEV]
Edge placement error (EPE) quantifies the distance between the edge of the target design and the edge of the actual printed feature on the wafer. When this distance exceeds a certain threshold, typically a few nanometers, 
the design is at risk of failing. Each instance where this threshold is surpassed is referred to as an EPE violation. A well-optimized mask should minimize the occurrence of these EPE violations as much as possible.
\end{mydefinition}

\begin{mydefinition}[PVB]
The process variation band (PVB) illustrates how the printed wafer image fluctuates due to variations in the manufacturing process. 
A common approach to quantitatively assess this variation involves perturbing simulation parameters related to system settings, such as the lens focus plane and UV dose strength. 
The area between the innermost and outermost contours of the printed image represents the PVB.
A smaller PVB indicates greater robustness of the mask against process variations.
\end{mydefinition}

We aim to incorporate additional metrics related to the mask shape itself. Although the definitive guidelines for curvilinear masks are still being developed \cite{choi2021curvilinear,pearman2020utilizing,armeanu2023application}, 
we introduce the isolated minimum shape area and minimum shape distance as key considerations for mask manufacturability. 
These metrics are commonly anticipated in the context of curvilinear mask design.

\begin{mydefinition}[MSA]
    The minimum shape area is the area in terms of $nm^2$ of the isolated shapes in the mask tile. Smaller isolated islands will easily cause process variations and mask manufacturing challenges. MSA is mathematically given by:
    \begin{align}
        \text{MSA} = \min \sum_{(i,j) \in \mathcal{S}_k} \vec{M}(i,j),~\forall k=0,1,...,N-1,
    \end{align}
    where $\mathcal{S}_k$ denotes the $k^{\text{th}}$ isolated shape in the mask $\vec{M}$ and $N$ is the total number of the isolated shapes.
\end{mydefinition}

\begin{mydefinition}[MSD]
    Similar to the minimum shape area (MSA), we want to avoid any two shapes being too close to each other. Therefore, we use the minimum shape distance to quantify this spacing.
    \begin{align}
        \text{MSD} = \min \text{dist} (\mathcal{S}_i, \mathcal{S}_j),~\forall i,j=0,1,...,N-1,
    \end{align}
    where
    \begin{align}
        \text{dist} (\mathcal{S}_i, \mathcal{S}_j) = \min \sqrt{(m-p)^2+(n-q)^2}, \\ \forall (m,n) \in \mathcal{S}_i, (p,q) \in \mathcal{S}_j. \nonumber
    \end{align}
\end{mydefinition}

\section{Algorithm}
\label{sec:algo}
This section outlines our GPU-accelerated curvilinear ILT algorithm. 
The key innovations include design retargeting and a differentiable morphological operator. 
These advancements are followed by the final ILT algorithm, which incorporates an enhanced optimization strategy compared to previous approaches.

\subsection{Design Retargeting}
\subsubsection{Lithography Is Low Pass Filter}
It is a well-established fact that the lithography process functions as a low-pass filter. This is mathematically represented by the complex lithography kernels $\vec{H}_i$ in \Cref{eq:aerial}.
Each entry in $\vec{H}_i$ corresponds to coefficients that affect the entire spectrum of the mask. However, the lithography kernel primarily captures information at locations associated with lower-frequency components of the mask. 
Consequently, high-frequency details, such as corners, are not effectively transferred to the silicon. 
As a result, optimizing a mask solely for Manhattan-shaped corners is not feasible and can lead to suboptimal performance at other critical locations, ultimately compromising the process windows.

\subsubsection{Curvilinear Design Retargeting}
In edge-based Optical Proximity Correction (OPC), a common industrial practice involves adjusting the placement of critical dimension error (EPE) measurement points to better guide the OPC process. 
Rather than positioning these measurement points directly on the polygon edge near corners—where they may lead to over-optimization—we instead relocate them either inside the shape at convex corners or outside the shape at concave corners. 
This adjustment helps the contour align more effectively with the target during OPC, thereby reducing the risk of over-optimization and enlarging process windows \cite{Calibre}.

\begin{figure}[tb!]
	\centering
	\subfloat[w/o CDR Gradient]{\pgfplotsset{
width =0.24\textwidth,
height=0.21\textwidth
}
\begin{tikzpicture}[scale=1]
\begin{axis}[minor tick num=0,
	xmin=0,
	xmax=200,
	ymax=20,
	ymin=0,
	xtick={0,40,80,120,160,200},
	yticklabel style={/pgf/number format/.cd, fixed, fixed zerofill, precision=2, /tikz/.cd},
	xlabel={Epoch},
	ylabel={Gradient},
	x label style={at={(axis description cs:0.5,-0.4)},anchor=south},
	y label style={at={(axis description cs:-0.1,1.1)},rotate=270,anchor=south},
	legend style={
		draw=none,
		at={(0.60,1.36)},
		anchor=north,
		legend columns=1,
	}
	]
	\addplot +[line width=0.7pt] [color=NVintel,   solid, mark=diamond, each nth point=5]  table [x={iter},  y={gradl2}]  {normal.dat};
	\addplot +[line width=0.7pt] [color=NVmgrey,   solid, mark=pentagon,each nth point=5]  table [x={iter},  y={gradcorner}]  {normal.dat};
	\addplot +[line width=0.7pt] [color=NVgreen,   solid, mark=triangle, each nth point=5]  table [x={iter},  y={gradpvb}]  {normal.dat};

\end{axis}
\end{tikzpicture} \label{fig:nocdr-gradient}}
	\subfloat[w/ CDR Gradient]{\pgfplotsset{
	width =0.24\textwidth,
	height=0.21\textwidth
}
\begin{tikzpicture}[scale=1]
	\begin{axis}[minor tick num=0,
		xmin=0,
		xmax=200,
		ymax=20,
		ymin=0,
		xtick={0,40,80,120,160,200},
		yticklabel style={/pgf/number format/.cd, fixed, fixed zerofill, precision=2, /tikz/.cd},
		xlabel={Epoch},
		ylabel={Gradient},
		x label style={at={(axis description cs:0.5,-0.4)},anchor=south},
		y label style={at={(axis description cs:-0.1,1.1)},rotate=270,anchor=south},
		legend style={
			draw=none,
			at={(0.60,1.36)},
			anchor=north,
			legend columns=1,
		}
		]
	\addplot +[line width=0.7pt] [color=NVintel,   solid, mark=diamond, each nth point=5]  table [x={iter},  y={gradl2}]  {retarget.dat};
\addplot +[line width=0.7pt] [color=NVmgrey,   solid, mark=pentagon,each nth point=5]  table [x={iter},  y={gradcorner}]  {retarget.dat};
\addplot +[line width=0.7pt] [color=NVgreen,   solid, mark=triangle, each nth point=5]  table [x={iter},  y={gradpvb}]  {retarget.dat};
		
		
	\end{axis}
\end{tikzpicture} \label{fig:cdr-gradient}} \\
	\subfloat[PVB Loss]{\pgfplotsset{
	width =0.24\textwidth,
	height=0.21\textwidth
}
\begin{tikzpicture}[scale=1]
	\begin{axis}[minor tick num=0,
		xmin=0,
		xmax=200,
		ymax=7500,
		ymin=6200,
		xtick={0,40,80,120,160,200},
		yticklabel style={/pgf/number format/.cd, fixed, fixed zerofill, precision=0, /tikz/.cd},
		xlabel={Epoch},
		ylabel={PVB},
		x label style={at={(axis description cs:0.5,-0.4)},anchor=south},
		y label style={at={(axis description cs:-0.1,1.1)},rotate=270,anchor=south},
		legend style={
			draw=none,
			at={(1.7,1)},
			anchor=north,
			legend columns=1,
		}
		]
		\addplot +[line width=0.7pt] [color=NVintel,   solid, mark=none, each nth point=1]  table [x={iter},  y={pvbn}]  {pvb.dat};
		\addplot +[line width=0.7pt] [color=NVgreen,   solid, mark=none,each nth point=1]  table [x={iter},  y={pvbr}]  {pvb.dat};
		\addplot +[line width=0.7pt] [color=NVintel,   solid, mark=diamond, each nth point=5]  {x};
		\addplot +[line width=0.7pt] [color=NVmgrey,   solid, mark=pentagon,each nth point=5]  {x};
		\addplot +[line width=0.7pt] [color=NVgreen,   solid, mark=triangle, each nth point=5]  {x};
		
		\legend{PVB w/o CDR,PVB w/ CDR, Grad-Full, Grad-Corner, Grad-PVB}
		
	\end{axis}
\end{tikzpicture} \label{fig:pvb}}
	\caption{Corner pixel mismatch generates over 20\% of the gradients during optimization. Design retargeting avoids over optimization on objectives that are unattainable.}
	\label{fig:cdr}
\end{figure}
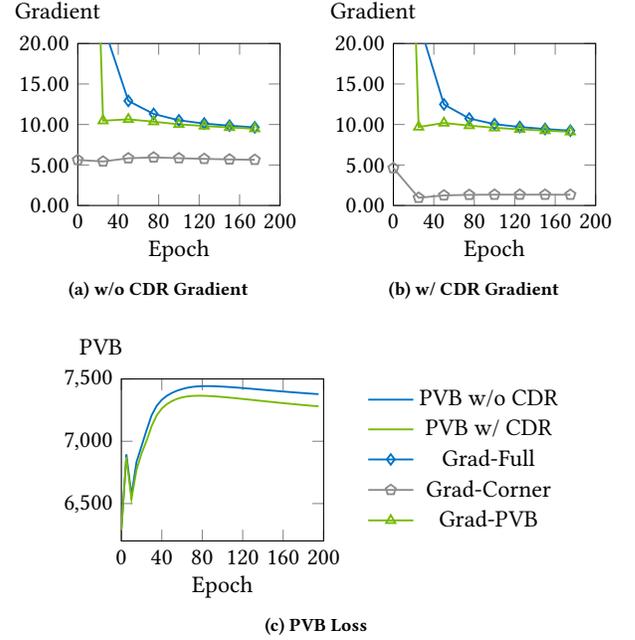

While the primary goal of inverse lithography technology (ILT) is to optimize the entire design image rather than just sampled control points, discrepancies between the resist and design can still result in significant gradients during optimization.
\Cref{fig:nocdr-gradient} illustrates the gradients produced by the objective functions change along the optimization process.
At a later training stage, we can observe that the gradients produced by the mismatch between polygon corner regions ($\approx$0.08\% of the total design area) still contribute 50\% of the overall gradient value. 
Consequently, the optimizer may focus excessively on objectives that are unattainable, leading to a negative impact on overall process variation. 
Additionally, the conventional approach of corner retargeting by moving EPE measurement points proves inadequate for pixel-based optimization objectives.
To address this issue, we introduce a method called curvilinear design retargeting (CDR). This technique smooths the original Manhattan design corners into curvilinear shapes, using the retargeted design as the objective for ILT optimization. 
The detailed algorithm is presented in \Cref{alg:cdr}, where we apply opening and closing operation on the original Manhattan design respectively (lines 3--4), followed by merging the morphological effects together (line 5).  
An example can be found in \Cref{fig:cdr-example}.
Note that CDR only modifies the vertex region of each polygon without touching the critical dimensions.

\begin{algorithm}[tb]
	\caption{CDR}
	\label{alg:cdr}
	\begin{algorithmic}[1]
		\Require Manhattan design $\vec{Z}^\ast$, convex corner smoothness coefficient $k_\text{cvx}$, concave corner smoothness coefficient $k_\text{ccv}$.
        \Ensure Corner retargeted design.
        \State $\vec{B}_\text{cvx} \gets$ Disc-shaped structuring element with size $k_\text{cvx}$;
        \State $\vec{B}_\text{ccv} \gets$ Disc-shaped structuring element with size $k_\text{ccv}$;
        \State $\vec{Z}^\ast_\text{cvx} \gets \vec{Z}^\ast \oplus \vec{B}_\text{cvx} \ominus \vec{B}_\text{cvx}$;
        \State $\vec{Z}^\ast_\text{ccv} \gets \vec{Z}^\ast \ominus \vec{B}_\text{ccv} \oplus \vec{B}_\text{ccv}$;
        \State $\vec{Z}^\ast \gets \vec{Z}^\ast_\text{cvx} + \vec{Z}^\ast_\text{ccv} - \vec{Z}^\ast$.
	\end{algorithmic}
\end{algorithm}

\subsection{Differentiable Morphological Operator}
Mask rules in ILT are typically addressed through post-processing \cite{OPC-DAC2023-Sun}, where small artifacts generated by ILT are manually removed. 
However, this approach inevitably leads to a loss of optimality, as will be demonstrated in the results section. 
To mitigate this issue, it is necessary to develop specific algorithms that can handle mask rule cleaning during the optimization process. 
Fortunately, we have observed that the properties of morphological operators align, to some extent, with the requirements of curvilinear mask rules. 
To leverage the benefits of morphological operators in ILT, these operators must be implemented in a differentiable manner.
We present the forward computing of dilation and erosion in \Cref{alg:dne}, which resembles regular convolution operation except that the summation over the sliding window is replaced with $\min$ (erosion) or $\max$ (dilation).
We follow the tradition in \texttt{Pytorch} to handle the gradients of $\min$ and $\max$.
The opening and closing can then be achieved following \Cref{eq:opening} and \Cref{eq:closing}.

\begin{algorithm}[tb]
	\caption{Differentiable Morphological Operator}
	\label{alg:dne}
	\begin{algorithmic}[1]
		\Function{Dilation\_Forward}{$\vec{A} \in \mathbb{R}^{N \times N},\vec{B} \in \mathbb{R}^{k \times k}$}
        \State $\vec{A}^{+} \gets \texttt{ZeroPadding}(\vec{A}, \floor{\frac{k}{2}})$;
        \For{each thread ${i,j} \in [0,N-1]$}
        \State $\vec{A}(i, j) = \max(\vec{A}^{+}(i:i+k,j:j+k) \odot \vec{B})$;
        \EndFor 
        \State \Return $\vec{A}$.
        \EndFunction
		\Function{Erosion\_Forward}{$\vec{A} \in \mathbb{R}^{N \times N},\vec{B} \in \mathbb{R}^{k \times k}$}
        \State $\vec{A}^{+} \gets \texttt{ZeroPadding}(\vec{A}, \floor{\frac{k}{2}})$;
        \For{each thread ${i,j} \in [0,N-1]$}
        \State $\vec{A}(i, j) = \min(\vec{A}^{+}(i:i+k,j:j+k) \odot \vec{B})$;
        \EndFor 
        \State \Return $\vec{A}$.
        \EndFunction
		\Function{Opening\_Forward}{$\vec{A} \in \mathbb{R}^{N \times N},\vec{B} \in \mathbb{R}^{k \times k}$}
        \State $\vec{A} \gets \texttt{Dilation\_Forward}(\texttt{Erosion\_Forward}(\vec{A}, \vec{B}), \vec{B})$;
        \State \Return $\vec{A}$.
        \EndFunction
		\Function{Closing\_Forward}{$\vec{A} \in \mathbb{R}^{N \times N},\vec{B} \in \mathbb{R}^{k \times k}$}
        \State $\vec{A} \gets \texttt{Erosion\_Forward}(\texttt{Dilation\_Forward}(\vec{A}, \vec{B}), \vec{B})$;      
        \State \Return $\vec{A}$.
        \EndFunction
	\end{algorithmic}
\end{algorithm}

\subsection{CurvyILT}
In this section, we will delve into our CurvyILT algorithm, detailing how it addresses the challenges of ILT and improves upon previous solutions. 
We will also discuss its expansion into a comprehensive full-chip solver.

\subsubsection{The Objectives}
The forward lithography process is well-established with precise mathematical formulations. The remaining components of the ILT solver involve the optimization objectives outlined in \Cref{eq:ilt-obj}. 
In line with conventional approaches, we specifically optimize the resist image mismatch under nominal conditions, along with the PVB area, with the former being closely linked to the EPE measurement.
We also introduce a term that implicitly improves the mask smoothness, noticing that rule-violating artifacts and notches generated by ILT algorithms are usually related to the perturbation in the high-frequency components of the mask.
Therefore, we define,
\begin{align}
    f(\vec{M},\vec{Z}^\ast) = &||\vec{Z}_\text{nom}-\vec{Z}^\ast||_2^2 + ||\vec{Z}_{\max}-\vec{Z}_{\min}||_2^2 \nonumber \\
    &+ \beta_3 ||\mathcal{F}(\vec{M})(k:;k:)||_2^2,
    \label{eq:curvyilt-obj}
\end{align}
where $\vec{Z}_\text{nom},\vec{Z}_{\min},\vec{Z}_{\max}$ are resist image computed through \Cref{eq:aerial,eq:cresist} under different process conditions (controlled by $\vec{\mathcal{H}}_i$'s),
$\mathcal{F}(\vec{M}_c)(k:;k:)$ is the Fourier transform of the continuous mask image dropping $k$ smallest frequency modes,
and $\beta_3$ is a manually determined parameter (at the scale of $1e-3$) that controls the mask's smoothness.
With the continuous and differentiable $f(\vec{M},\vec{Z}^\ast)$, we are able to solve ILT through a gradient-based approach. 
We use \texttt{Adam} optimizer for better convergence. 

\subsubsection{CurvyILT Solver}
The detailed CurvyILT algorithm is elaborated in \Cref{alg:curvyilt}, which can be viewed in three phases.
The first phase is design preprocessing including curvilinear design retargeting (line 1) and mask initialization (lines 2--3).
The second phase covers the major optimization steps and we update the mask iteratively till reach the maximum optimization steps (lines 5--16).
During each optimization step, we heuristically apply morphological operators on the mask image to perform corner smoothing (lines 8--10) and remove artifacts (lines 11--12).
We employ the Adam optimizer to compute the true mask gradients and mask update steps for best practices (lines 13--16).
The last phase can be deemed as post-processing where the mask is scaled back to the desired resolution through interpolation and binarized with a preset threshold (lines 17--19).

\begin{algorithm}[tb]
	\caption{CurvyILT}
	\label{alg:curvyilt}
	\begin{algorithmic}[1]
		\Require Rasterized design target $\vec{Z}^\ast \in \{0,1\}$, maximum optimization steps $T$, mask steepness $\beta_1$, resist steepness $\beta_2$, mask smoothness $\beta_3$, learning rate $\lambda$, $k_\text{cvx}$, $k_\text{ccv}$, mask resolution scaling factor $s$, morphological operator kernel $k_\text{morph}$, mask binarization threshold $M_s$, morphological gradient step $t_\text{morph\_step}$, iteration number morphological gradient taking effect $t_\text{morph}$;
		\Ensure Optimized mask $\vec{M}$;
        \State $\vec{Z}^\ast_r \gets$ \texttt{CDR}($\vec{Z}^\ast$,$k_\text{cvx}$,$k_\text{ccv}$); \Comment{\Cref{alg:cdr}}
		\State $\vec{Z}_s^\ast \gets$ Design target $\vec{Z}_r^\ast$ downsample by $s$; \Comment{Adjust rasterized design resolution to balance speed and performance}
		\State $\vec{M} \gets \vec{Z}^\ast_s$; \Comment{Mask initialized as design target}
        \State $\vec{B}_\text{morph} \gets$ Disc-shaped structuring elements with size $k_\text{morph}$;
        \State $\vec{C}_\text{morph} \gets$ Disc-shaped structuring elements with size $s \times k_\text{morph}$;
        \For{t=1,2,...,T}
        \State $\vec{M}_c \gets \dfrac{1}{1+\exp[-\beta_1 (\vec{M}-M_s)]}$;
        \If{$t>t_\text{morph}~\text{and}~t \% t_\text{morph\_step}=0$} \Comment{Cleaning mask artifacts through differentiable morphological operator.}
        \State $\vec{M}_\text{open} \gets \texttt{Opening\_Forward}(\vec{M}_c, \vec{B}_\text{morph})$;
        \State $\vec{M}_\text{close} \gets \texttt{Closing\_Forward}(\vec{M}_c, \vec{B}_\text{morph})$;
        \State $\vec{M}_c \gets \vec{M}_\text{open} + \vec{M}_\text{close} - \vec{M}_c$;
        \EndIf
        \State $\vec{Z}_\text{nom}, \vec{Z}_\text{max}, \vec{Z}_\text{min} \gets \texttt{LithoSim}(\vec{M}_c)$; \Comment{\cref{eq:aerial,eq:cresist}.}
        \State $f(\vec{M}_c,\vec{Z}^\ast_s) \gets$ Compute the current loss w.r.t. the current mask;
        \State $\vec{G} \gets \texttt{AdamOpt}(\nabla_f \vec{M}, \lambda)$; \Comment{Compute the actual gradient step to update the mask, we discovered Adam optimizer brings best optimization quality.}
        \State $\vec{M} \gets \vec{M} - \vec{G}$;
        \EndFor
        \State $\vec{M} \gets \texttt{Interpolation}(\vec{M}, \text{scale\_factor}=s, \text{anti\_alias}=\textbf{True}, \text{mode}=\text{``bicubic''})$; \Comment{Scale the mask back to desired resolution.}
        \State $\vec{M}(i,j) \gets 1, \forall \vec{M}(i,j)>M_s$; \Comment{Mask processing.}
        \State $\vec{M}(i,j) \gets 0, \forall \vec{M}(i,j) \leq M_s$;
        \State $\vec{M}_\text{open} \gets \texttt{Opening\_Forward}(\vec{M}, \vec{C}_\text{morph})$;
        \State $\vec{M}_\text{close} \gets \texttt{Closing\_Forward}(\vec{M}, \vec{C}_\text{morph})$;
        \State $\vec{M} \gets \vec{M}_\text{open} + \vec{M}_\text{close} - \vec{M}_c$;
        \State $\vec{M} \gets \texttt{Opening\_Forward}(\vec{M}, \vec{C}_\text{morph})$;
        \State $\vec{M} \gets \texttt{Closing\_Forward}(\vec{M}, \vec{C}_\text{morph})$.
	\end{algorithmic}
\end{algorithm}

\subsection{Discussion}
Compared to previous methods, the proposed algorithm offers \textbf{faster convergence} by utilizing curvilinear design retargeting and delivers \textbf{improved mask quality} through the application of differentiable morphological operators in accordance with mask design rules.
Additionally, by addressing ILT-generated artifacts during the optimization process, our approach eliminates the need for unnecessary post-processing, preserving the optimized resist image quality.
While earlier efforts have employed strided average pooling to smooth the mask shape, this occurs after each iteration’s mask binarization step, leading to inaccuracies in the forward lithography calculations.

\begin{table}[]
\centering
\caption{Benchmark Statistics.}
\label{tab:bench}
\def\arraystretch{1.2}
\setlength{\tabcolsep}{10pt}
\begin{tabular}{c|c|c}
\toprule
Benchmark                   & Layer & Statistic                                   \\ \midrule
ICCAD13                     & Metal & 10                                          \\  \midrule
\multirow{2}{*}{LithoBench} & Metal & 271 \\ 
                            & Via   & 165       \\ \bottomrule
\end{tabular}
\end{table}

\begin{table*}[]
\centering
\caption{Result comparison on \iccad~with state-of-the-art ILT solvers.}
\label{tab:result-iccad13}
\def\arraystretch{1.2}
\setlength{\tabcolsep}{2pt}
\begin{tabular}{c|ccc|ccc|ccccc|ccccc}
\toprule
\multirow{2}{*}{Case} & \multicolumn{3}{c|}{A2-ILT \cite{OPC-DAC2022-Wang} } & \multicolumn{3}{c|}{MultiILT \cite{OPC-DAC2023-Sun} Ref} & \multicolumn{5}{c|}{MultiILT \cite{OPC-DAC2023-Sun} Reimp} & \multicolumn{5}{c}{Ours}                                                             \\
                            & MSE                    & PV                     & EPE                & MSE                      & PV                       & EPE                 & MSE             & PV              & EPE         & MSA          & MSD        & MSE              & PV               & EPE          & MSA             & MSD           \\ \midrule
1                           & 45824                  & 59136                  & 7                  & 38495                    & 47015                    & 3                   & 39533           & 44887           & 3           & 832          & 1          & 38066            & 44447            & 3            & 1062            & 16            \\
2                           & 33976                  & 52054                  & 3                  & 28173                    & 37555                    & 0                   & 32516           & 37374           & 0           & 640          & 12         & 28623            & 36914            & 0            & 2029            & 28            \\
3                           & 94634                  & 82661                  & 62                 & 67949                    & 69361                    & 22                  & 65315           & 75011           & 23          & 768          & 9          & 61650            & 70580            & 15           & 981             & 10            \\
4                           & 20405                  & 29435                  & 2                  & 10307                    & 21514                    & 0                   & 9099            & 21484           & 0           & 704          & 1          & 9211             & 21584            & 0            & 1342            & 10            \\
5                           & 37038                  & 62068                  & 1                  & 28482                    & 49683                    & 0                   & 30015           & 48696           & 0           & 896          & 9          & 27859            & 47870            & 0            & 865             & 11            \\
6                           & 40701                  & 54842                  & 2                  & 30334                    & 44127                    & 0                   & 33400           & 42788           & 0           & 896          & 9          & 30391            & 42288            & 0            & 1088            & 17            \\
7                           & 21840                  & 48474                  & 0                  & 14635                    & 36961                    & 0                   & 17419           & 36241           & 0           & 768          & 9          & 12791            & 34389            & 0            & 2061            & 12            \\
8                           & 14912                  & 24598                  & 0                  & 11194                    & 20985                    & 0                   & 11552           & 18987           & 0           & 640          & 9          & 11468            & 18649            & 0            & 621             & 9             \\
9                           & 47489                  & 68056                  & 2                  & 34900                    & 54948                    & 0                   & 37219           & 54792           & 0           & 640          & 1          & 32720            & 54387            & 0            & 3940            & 13            \\
10                          & 9399                   & 20243                  & 0                  & 7266                     & 16581                    & 0                   & 7180            & 14979           & 0           & 2304         & 9          & 7130             & 15014            & 0            & 2964            & 68            \\ \midrule
Avg                     & 36621.8                & 50156.7                & 7.9                & 27173.5                  & 39873.0                  & 2.5                 & 28324.8         & 39523.9         & 2.6         & 908.8        & 6.9        & \textbf{25990.9} & \textbf{38612.2} & \textbf{1.8} & \textbf{1695.3} & \textbf{19.4} \\ \bottomrule
\end{tabular}
\end{table*}

\begin{table}[]
\centering
\caption{Result comparison on \lithobench~with state-of-the-art ILT solvers.}
\label{tab:result-lithobench}
\def\arraystretch{1.2}
\setlength{\tabcolsep}{1pt}
\begin{tabular}{c|ccccc|ccccc}
\toprule
\multirow{2}{*}{Case} & \multicolumn{5}{c|}{MultiILT \cite{OPC-DAC2023-Sun} Ref}    & \multicolumn{5}{c}{Ours}      \\
                      & MSE     & PV      & EPE  & MSA   & MSD  & MSE     & PV      & EPE & MSA    & MSD  \\ \midrule
Metal                     & 13814.2 & 24928.2 & 0.03 & 720.4 & 16.1 & 14643.3 & 21631.3 & 0.0 & 1709.0 & 24.4 \\
Via                    & 34813.4 & 39997.4 & 8.6  & 464.2 & 19.2 & 29183.8 & 36172.4 & 3.8 & 1072.9 & 11.3 \\ \midrule
Avg                   & 24313.8 & 32462.8 & 4.31 & 592.3 & 17.6 & 21913.5 & 28901.9 & 1.9 & 1390.9 & 17.8 \\ \bottomrule
\end{tabular}
\end{table}

\begin{table}[]
\centering
\caption{Efficiency of state-of-the-art ILT solvers.}
\label{tab:effi-iccad13}
\def\arraystretch{1.2}
\setlength{\tabcolsep}{2pt}
\begin{tabular}{c|ccc}
\toprule
Solver       & A2-ILT \cite{OPC-DAC2022-Wang} & MultiILT \cite{OPC-DAC2023-Sun} & Ours   \\ \midrule
Throughput & 4.51   & 3.45     & 2.11   \\
OptPeakMemory   & -      & 7.2GB    & 0.6GB \\ \bottomrule
\end{tabular}
\end{table}

\begin{table}[]
\centering
\caption{Post processing removes rule-violating artifacts at the cost of optimality.}
\label{tab:pp}
\def\arraystretch{1.2}
\setlength{\tabcolsep}{1pt}
\begin{tabular}{c|ccccc|ccccc}
\toprule
\multirow{2}{*}{Case} & \multicolumn{5}{c|}{MultiILT w/o PP} & \multicolumn{5}{c}{MultiILT w/ PP}              \\
                            & MSE             & PV            & EPE         & MSA           & MSD         & MSE     & PV      & EPE & MSA   & MSD \\ \midrule
1                           & 37976           & 45423         & 3           & 64            & 1           & 39533   & 44887   & 3   & 832   & 1   \\
2                           & 31070           & 37754         & 0           & 576           & 12          & 32516   & 37374   & 0   & 640   & 12  \\
3                           & 63036           & 73396         & 19          & 64            & 1           & 65315   & 75011   & 23  & 768   & 9   \\
4                           & 8498            & 21561         & 0           & 64            & 1           & 9099    & 21484   & 0   & 704   & 1   \\
5                           & 28478           & 49400         & 0           & 64            & 9           & 30015   & 48696   & 0   & 896   & 9   \\
6                           & 29666           & 43162         & 0           & 64            & 1           & 33400   & 42788   & 0   & 896   & 9   \\
7                           & 17333           & 36319         & 0           & 256           & 9           & 17419   & 36241   & 0   & 768   & 9   \\
8                           & 11486           & 19048         & 0           & 64            & 1           & 11552   & 18987   & 0   & 640   & 9   \\
9                           & 34459           & 55688         & 0           & 64            & 1           & 37219   & 54792   & 0   & 640   & 1   \\
10                          & 7180            & 14979         & 0           & 2304          & 9           & 7180    & 14979   & 0   & 2304  & 9   \\ \midrule
Avg                     & 26918.2         & 39673         & 2.2         & 358.4         & 4.5         & 28324.8 & 39523.9 & 2.6 & 908.8 & 6.9 \\ \bottomrule
\end{tabular}
\end{table}

\section{Experiments}
\label{sec:result}
\subsection{Dataset and Configurations}
To assess the algorithm's performance, we utilize the widely recognized benchmark suites from the ICCAD13 CAD Contest (\texttt{ICCAD13}) \cite{OPC-ICCAD2013-Banerjee} and the more recent \texttt{LithoBench} dataset \cite{lithobench}. The details of these benchmarks are summarized in \Cref{tab:bench}. 
The \iccad~benchmark consists of ten $2\mu m \times 2\mu m$ clips featuring M1 layer polygons. 
Meanwhile, \lithobench~is a larger dataset originally developed for AI applications, comprising over 10,000 metal and via layer clips. For demonstration purposes, we focused on the standard cell collection within \lithobench, which includes 271 metal layer designs and 165 via layer designs.
We implement the algorithm using \texttt{Pytorch} with CUDA support and adopt the EPE checker from NeuralILT \cite{OPC-ICCAD2020-NeuralILT}.
To compare our approach with prior arts, we also implemented the multi-level ILT \cite{OPC-DAC2023-Sun} that reproduces the original results as closely as possible.
All experiments are conducted on a single NVIDIA RTX A6000 platform with 48GB memory.

\subsection{Comparison with State-of-the-art}
In this first experiment, we compare our method with state-of-the-art ILT solvers on ten \iccad~clips as shown in \Cref{tab:result-iccad13}, where
columns ``MSE'', ``PV'', ``EPE'', ``MSA'', and ``MSD'' denote the mean square error between the resist image and the design, PVBand area, EPE violation count, minimum shape area ($nm^2$) and minimum shape distance ($nm$), respectively;
column ``A2-ILT \cite{OPC-DAC2022-Wang}'' lists the results of ILT solve with the gradient regularization of mask image attention map;
column ``MultiILT \cite{OPC-DAC2023-Sun} Ref'' corresponds to the results of the SOTA academic GPU-accelerated ILT solver that takes advantage of lithography simulation at different rasterization resolutions;
column ``MultiILT \cite{OPC-DAC2023-Sun} Reimp'' corresponds to the author re-implemented MultiILT (exact) as we need to reproduce the mask results to measure certain mask rules and perform additional studies;
column ``Ours'' lists our full algorithm implementation results with all techniques enabled. 

We performed a comprehensive parameter search of our implementation on MultiILT to align the reported results, with only a minor discrepancy observed in case 3, which does not impact our overall conclusions. Our findings demonstrate that our method significantly outperforms existing approaches in terms of MSE, PVB, and EPE. 
Notably, in case 3—a particularly complex high-density clip—our solution achieved a reduction in EPE violations to below 20 for the first time. On average, our method lowers the EPE violation count by 30\% and the resist MSE by 8.6\%, while maintaining a favorable process window.
Additionally, our approach exhibits notable improvements in curvilinear mask complexity, offering larger minimum shape areas and distances, thus enhancing its potential for curvilinear ILT solver development.
It is important to mention that the MultiILT results are subject to post-processing, where minor artifacts are manually removed, which may affect optimality. 
\Cref{tab:result-lithobench} also lists the optimization result on the standard cell collection in \lithobench.
Similarly, on more practical design layers, our solver outperforms prior work by an even larger amount with $2\times$ smaller EPE violation count and 11\% PV area reduction.

We have adjusted our post-processing configurations to ensure that the final quality of results for MultiILT \cite{OPC-DAC2023-Sun} aligns closely with the values reported in the original manuscript.
Furthermore, we compare the efficiency of different solvers in \Cref{tab:effi-iccad13}, with the first row presenting throughput (seconds per clip) for three recent ILT solvers.
Our algorithm demonstrates superior efficiency in terms of both runtime and peak memory usage during the optimization runtime.
Notably, MultiILT requires ten times more GPU memory than our approach due to its memory-intensive high-resolution phase, which our algorithm effectively avoids.

\subsection{Post Processing Sabotages Optimality}
In the second experiment, we investigate the effects of post processing in MultiILT \cite{OPC-DAC2023-Sun} and demonstrate the necessity of mask cleansing during optimization runtime. 
As shown in \Cref{tab:pp}, 
column ``MultiILT w/o PP'' lists the optimization results without post-processing,
and column ``MultiILT w/ PP'' corresponds to the optimization results with post-processing.
We observe a substantial degradation in results when ILT-generated artifacts are removed, highlighting a critical trade-off between mask complexity and ILT quality of results (QoR). In contrast, our approach manages mask complexity throughout the optimization process and eliminates the need for post-processing, thereby preserving optimality.

\begin{table*}[]
\centering
\caption{Ablation study.}
\label{tab:ablation}
\def\arraystretch{1.1}
\setlength{\tabcolsep}{1pt}
\begin{tabular}{c|ccccc|ccccc|ccccc|ccccc}
\toprule
\multirow{2}{*}{Case} & \multicolumn{5}{c|}{Ours (Baseline)}              & \multicolumn{5}{c|}{Ours (CDR)}              & \multicolumn{5}{c|}{Ours (Morph)}               & \multicolumn{5}{c}{Ours (CDR+Morph)}                \\
                      & MSE     & PV      & EPE & MSA   & MSD  & MSE     & PV      & EPE & MSA   & MSD  & MSE     & PV      & EPE & MSA    & MSD  & MSE     & PV      & EPE & MSA    & MSD  \\ \midrule
1                     & 35938   & 44969   & 3   & 1049  & 10   & 37077   & 44355   & 3   & 101   & 11   & 36583   & 44975   & 3   & 587    & 24   & 37783   & 44446   & 3   & 1062   & 16   \\
2                     & 27796   & 37311   & 0   & 504   & 14   & 28996   & 36959   & 0   & 28    & 25   & 27661   & 37314   & 0   & 2101   & 19   & 28644   & 36940   & 0   & 2030   & 28   \\
3                     & 62410   & 72333   & 15  & 320   & 6    & 60599   & 70559   & 15  & 82    & 12   & 60124   & 72046   & 14  & 999    & \textcolor{red}{3}    & 62212   & 70545   & 15  & 979    & 12   \\
4                     & 8334    & 21801   & 0   & 8     & \textcolor{red}{2}    & 8958    & 21446   & 0   & 358   & 11   & 8496    & 21957   & 0   & 1912   & 20   & 9193    & 21577   & 0   & 1346   & 10   \\
5                     & 27121   & 48497   & 0   & 508   & 8    & 27831   & 47911   & 0   & 85    & 13   & 27116   & 48453   & 0   & 547    & 8    & 27824   & 47861   & 0   & 841    & 11   \\
6                     & 29858   & 42777   & 0   & 1746  & 12   & 30243   & 42277   & 0   & 50    & 26   & 29970   & 42826   & 0   & 1484   & 19   & 30391   & 42287   & 0   & 1088   & 17   \\
7                     & 12388   & 34568   & 0   & 628   & 22   & 12603   & 34253   & 0   & 465   & 23   & 12638   & 34809   & 0   & 1384   & \textcolor{red}{1}    & 12801   & 34409   & 0   & 2060   & 12   \\
8                     & 10963   & 18838   & 0   & 1034  & 26   & 11343   & 18572   & 0   & 484   & 30   & 11087   & 18921   & 0   & 789    & 23   & 11473   & 18644   & 0   & 656    & 9    \\
9                     & 31695   & 55250   & 0   & 98    & 16   & 32723   & 54399   & 0   & 36    & 12   & 31830   & 55296   & 0   & 843    & 12   & 32723   & 54393   & 0   & 3939   & 13   \\
10                    & 6743    & 15097   & 0   & 3217  & 66   & 7054    & 14951   & 0   & 2797  & 68   & 6864    & 15288   & 0   & 3750   & 66   & 7127    & 15013   & 0   & 2981   & 68   \\ \midrule
Avg                   & 25324.6 & 39144.1 & 1.8 & 911.2 & 18.2 & 25742.7 & 38568.2 & 1.8 & 448.6 & 23.1 & 25236.9 & 39188.5 & 1.7 & 1439.6 & 19.5 & 26017.1 & 38611.5 & 1.8 & 1698.2 & 19.6 \\ \bottomrule
\end{tabular}
\end{table*}
\subsection{Ablation Study}
In this experiment, we will demonstrate the effectiveness of each techniques proposed in this paper through ablation study. 
Again, we use the ten \iccad~clips as example for simplicity.
We show the ILT results change with different techniques being included gradually in \Cref{tab:ablation}. 
Compared to the baseline implementation, CDR avoids excessive optimization on polygon corners, encourages sub-resolution assist feature (SRAF) growth and brings better process window (1.5\% PV reduction) while inducing small artifacts that challenges mask manufacturing (51\% MSA degradation).

Morphological operators enhance the ILT process in several key ways: 1) They eliminate small artifacts generated during ILT, resulting in a final mask with a larger minimum shape area (58\% improvement in MSA). 2) The frequent removal of small shapes during optimization provides regularization and introduces stochastic effects, leading to a slight improvement in EPE violations in case 3, although this benefit is marginal.

While morphological operators can impose penalties on minimum shape distance—due to the sequential closing after opening in our algorithm, which prevents nearby shapes from merging—this penalty can be effectively mitigated through design retargeting, thanks to the SRAF growing effect. With all techniques applied, our approach achieves significantly improved MSA and MSD without compromising performance.

\begin{figure*}[tb!]
	\centering
	\subfloat[Epoch 10]{\includegraphics[width=.7\textwidth]{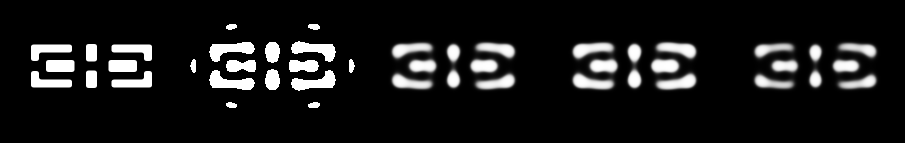}} \\
     \subfloat[Epoch 30]{\includegraphics[width=.7\textwidth]{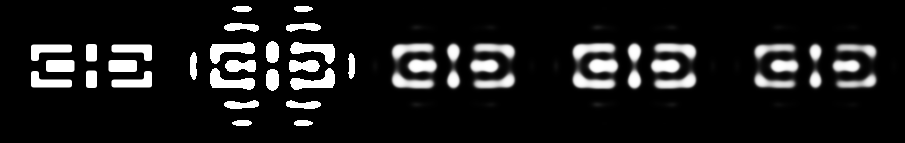}} \\
     \subfloat[Epoch 50]{\includegraphics[width=.7\textwidth]{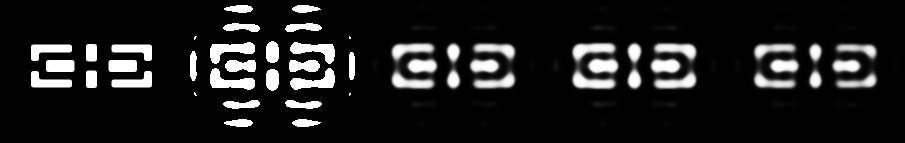} \label{fig:e50}} \\
     \subfloat[Epoch 70]{\includegraphics[width=.7\textwidth]{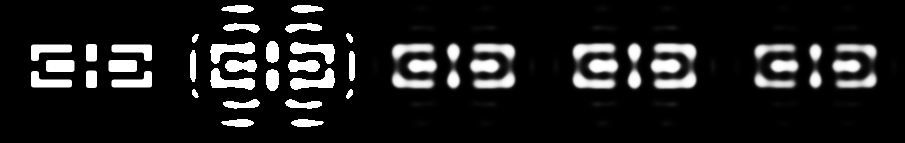} \label{fig:e70}} \\
     \subfloat[Epoch Final]{\includegraphics[width=.7\textwidth]{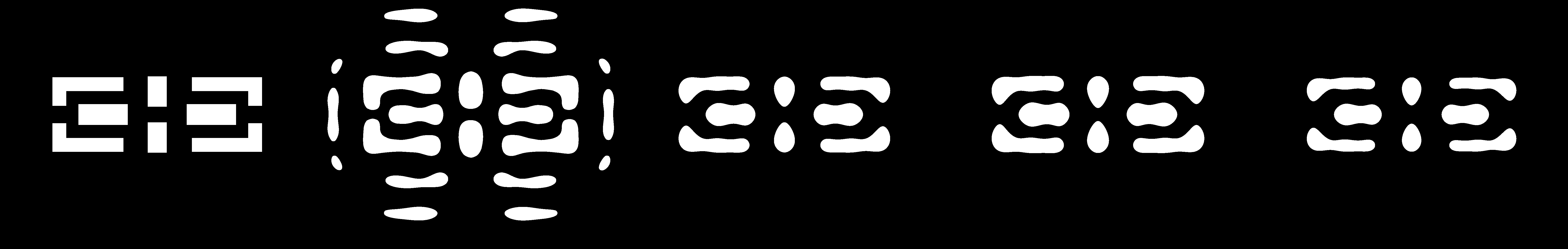} \label{fig:efinal}}
	\caption{Visualization of the optimization trajectory of our algorithm. From left to right: CDR design, mask, nominal condition image, outermost image, and innermost image.}
	\label{fig:opt}
\end{figure*}
\subsection{Result Visualization}
Finally, we will visualize some ILT results to illustrate the advantages of our algorithm more clearly.
Here, we use a simple clip from \iccad~ as an example and depict the optimization trajectory.
We can observe at a later optimization stage (\Cref{fig:e50}), the morphological operators start to take effect and try to remove notches generated on the mask edge and enlarge thin mask critical dimensions. 
This also ensures a minimum change of the optimization status when producing the final smoothed high-resolution masks (\Cref{fig:efinal}).

\section{Conclusion}
\label{sec:conclu}
In this paper, we explored GPU-accelerated ILT algorithms for curvilinear mask generation, assessing current pixel-based mask optimization methods and identifying key challenges in enhancing mask optimization for future multi-beam mask makers. 
We introduced innovative techniques to reduce mask complexity and enhance printability, such as curvilinear design retargeting, the incorporation of differentiable morphological operators in optimization, and the inclusion of mask smoothing objectives. 
These elements converge to form the curvyILT algorithm, which demonstrates superior performance against leading academic solvers across various pattern complexities. For practical application and deployment, future research will focus on covering more curvilinear mask rules, scaling to full-chip designs, and integrating AI technologies. 
To encourage further innovation in computational lithography and semiconductor manufacturing, we will make our source code publicly available.

\clearpage
{
    \bibliographystyle{IEEEtran}
    \bibliography{ref/Top,ref/DFM,ref/Additional,ref/LLM,ref/PD,ref/HSD}

\begin{thebibliography}{10}
\providecommand{\url}[1]{#1}
\csname url@samestyle\endcsname
\providecommand{\newblock}{\relax}
\providecommand{\bibinfo}[2]{#2}
\providecommand{\BIBentrySTDinterwordspacing}{\spaceskip=0pt\relax}
\providecommand{\BIBentryALTinterwordstretchfactor}{4}
\providecommand{\BIBentryALTinterwordspacing}{\spaceskip=\fontdimen2\font plus
\BIBentryALTinterwordstretchfactor\fontdimen3\font minus
  \fontdimen4\font\relax}
\providecommand{\BIBforeignlanguage}[2]{{%
\expandafter\ifx\csname l@#1\endcsname\relax
\typeout{** WARNING: IEEEtran.bst: No hyphenation pattern has been}%
\typeout{** loaded for the language `#1'. Using the pattern for}%
\typeout{** the default language instead.}%
\else
\language=\csname l@#1\endcsname
\fi
#2}}
\providecommand{\BIBdecl}{\relax}
\BIBdecl

\bibitem{OPC-DATE2015-Kuang}
J.~Kuang, W.-K. Chow, and E.~F.~Y. Young, ``A robust approach for process
  variation aware mask optimization,'' in \emph{IEEE/ACM Proceedings Design,
  Automation and Test in Eurpoe (DATE)}, 2015, pp. 1591--1594.

\bibitem{OPC-JM3-2016-Matsunawa}
T.~Matsunawa, B.~Yu, and D.~Z. Pan, ``Optical proximity correction with
  hierarchical bayes model,'' \emph{Journal of Micro/Nanolithography, MEMS, and
  MOEMS (JM3)}, vol.~15, no.~2, p. 021009, 2016.

\bibitem{MEEF-TSM2000-Wong}
A.~K. Wong, R.~A. Ferguson, and S.~M. Mansfield, ``The mask error factor in
  optical lithography,'' \emph{IEEE Transactions on Semiconductor Manufacturing
  (TSM)}, vol.~13, no.~2, pp. 235--242, 2000.

\bibitem{mbmw}
H.~Matsumoto, J.~Yasuda, T.~Motosugi, H.~Kimura, M.~Kawaguchi, Y.~Kojima,
  H.~Yamashita, M.~Saito, T.~Tamura, and N.~Nakayamada, ``Multi-beam mask
  writer mbm-3000 for next generation euv mask production,'' in \emph{Photomask
  Technology 2023}, vol. 12751.\hskip 1em plus 0.5em minus 0.4em\relax SPIE,
  2023.

\bibitem{curvyMRC}
J.~Sturtevant, ``Curves ahead! ic manufacturing prepares for curvilinear
  masks,'' Siemens, Tech. Rep., 2023.

\bibitem{OPC-DAC2022-Wang}
Q.~Wang, B.~Jiang, M.~D. Wong, and E.~F. Young, ``{A2-ILT}: {GPU} accelerated
  {ILT} with spatial attention mechanism,'' in \emph{ACM/IEEE Design Automation
  Conference (DAC)}, 2022.

\bibitem{OPC-DATE2021-Yu}
Z.~Yu, G.~Chen, Y.~Ma, and B.~Yu, ``A {GPU}-enabled level set method for mask
  optimization,'' in \emph{IEEE/ACM Proceedings Design, Automation and Test in
  Eurpoe (DATE)}, 2021.

\bibitem{OPC-DAC2023-Sun}
S.~Sun, F.~Yang, B.~Yu, L.~Shang, and X.~Zeng, ``Efficient ilt via multi-level
  lithography simulation,'' in \emph{ACM/IEEE Design Automation Conference
  (DAC)}, 2023, pp. 1--6.

\bibitem{openilt}
S.~Zheng, B.~Yu, and M.~Wong, ``Openilt: An open source inverse lithography
  technique framework,'' in \emph{IEEE International Conference on ASIC
  (ASICON)}, 2023, pp. 1--4.

\bibitem{choi2021curvilinear}
Y.~Choi, A.~Fujimura, and A.~Shendre, ``Curvilinear masks: an overview,''
  \emph{Proceedings of SPIE}, vol. 11855, pp. 157--172, 2021.

\bibitem{pearman2020utilizing}
R.~Pearman, D.~O'Riordan, J.~Ungar, M.~Niewczas, L.~Pang, and A.~Fujimura,
  ``How utilizing curvilinear design enables better manufacturing process
  window,'' in \emph{Proceedings of SPIE}, vol. 11328.\hskip 1em plus 0.5em
  minus 0.4em\relax SPIE, 2020, pp. 183--191.

\bibitem{armeanu2023application}
A.-M. Armeanu, E.~Malankin, N.~Lafferty, C.-I. Wei, M.~K. Sears, G.~Fenger,
  X.~Zhang, W.~Gillijns, D.~Trivkovic, R.-h. Kim \emph{et~al.}, ``Application
  of resolution enhancement techniques at high na euv for next generation dram
  patterning,'' in \emph{Proceedings of SPIE}, vol. 12495.\hskip 1em plus 0.5em
  minus 0.4em\relax SPIE, 2023, pp. 52--63.

\bibitem{Calibre}
``{Calibre},'' \url{https://eda.sw.siemens.com/en-US/ic/calibre-design/}.

\bibitem{morphological}
M.~L. Comer and E.~J. Delp, \emph{Morphological operations}.\hskip 1em plus
  0.5em minus 0.4em\relax Boston, MA: Springer US, 1998, pp. 210--227.

\end{thebibliography}
}

\end{document}